\documentclass[10pt,twocolumn,letterpaper]{article}

\usepackage{cvpr}
\usepackage{times}
\usepackage{epsfig}
\usepackage{graphicx}
\usepackage{amsmath}
\usepackage{amssymb}
\usepackage{stmaryrd}
\usepackage{soul}
\usepackage{array}
\usepackage{multirow}
\usepackage{hhline}
\usepackage{subfigure}
\usepackage{dblfloatfix}

\newcommand{\todo}[1]{}
\newcommand{\note}[1]{}
\newcommand{\zh}[1]{}

\newcommand{\B}[1]{\mathbf{#1}}

\newcommand{\X}[1]{\textbf{#1}}

\newcommand\Tstrut{\rule{0pt}{2.6ex}}         
\newcommand\Bstrut{\rule[-0.9ex]{0pt}{0pt}}   

\usepackage[pagebackref=true,breaklinks=true,letterpaper=true,colorlinks,bookmarks=false]{hyperref}

\cvprfinalcopy 

\def\cvprPaperID{2429} 

\begin{document}

\title{Boundary-aware Instance Segmentation} 

\author{Zeeshan Hayder$^{1,2}$, Xuming He$^{2,1}$\\
$^1$Australian National University \& $^2$Data61/CSIRO
\thanks{Data61/CSIRO is funded by the Australian Government as represented by the Department of Broadband, Communications and the Digital Economy and
the ARC through the ICT Centre of Excellence program.}
\and
Mathieu Salzmann$^{3}$\\
$^3$CVLab, EPFL, Switzerland\\
}

\maketitle

\begin{abstract}
We address the problem of instance-level semantic segmentation, which aims at jointly detecting, segmenting and classifying every individual object in an image. 
In this context, existing methods typically propose candidate objects, usually as bounding boxes, and directly predict a binary mask within each such proposal. As a consequence, they cannot recover from errors in the object candidate generation process, such as too small or shifted boxes.

In this paper, we introduce a novel object segment representation based on the distance transform of the object masks. We then design an object mask network (OMN) with a new residual-deconvolution architecture that infers such a representation and decodes it into the final binary object mask. This allows us to predict masks that go beyond the scope of the bounding boxes and are thus robust to inaccurate object candidates.
We integrate our OMN into a Multitask Network Cascade framework, and learn the resulting boundary-aware instance segmentation (BAIS) network in an end-to-end manner. 
Our experiments on the PASCAL VOC 2012 and the Cityscapes datasets demonstrate the benefits of our approach, which outperforms the state-of-the-art in both object proposal generation and instance segmentation.
\end{abstract}

\section{Introduction}

Instance-level semantic segmentation, which aims at jointly detecting, segmenting and classifying every individual object in an image, has recently become a core challenge in scene understanding~\cite{Cordts2016Cityscapes,DBLP:COCO14,Everingham12}. Unlike its category-level counterpart, instance segmentation provides detailed information about the location, shape and number of individual objects. As such, it has many applications in diverse areas, such as autonomous driving~\cite{Ziyu15}, personal robotics~\cite{gupta2014learning} and plant analytics~\cite{scharr2014annotated}.

Existing approaches to multiclass instance segmentation typically rely on generic object proposals in the form of bounding boxes. These proposals can be learned~\cite{HariharanAGM14,LiHM15,Jifeng15cocoseg} or sampled by sliding windows~\cite{DeepMask,Jifeng16cocoseg}, and greatly facilitate the task of identifying the different instances, may they be from the same category or different ones. Object segmentation is then achieved by predicting a binary mask within each box proposal, which can then be classified into a semantic category. This approach to segmentation, however, makes these methods sensitive to the quality of the bounding boxes; they cannot recover from errors in the object proposal generation process, such as too small or shifted boxes.

\begin{figure}[t]
	\centering
	\includegraphics[trim = 0 170 530 0, clip, width=\columnwidth]{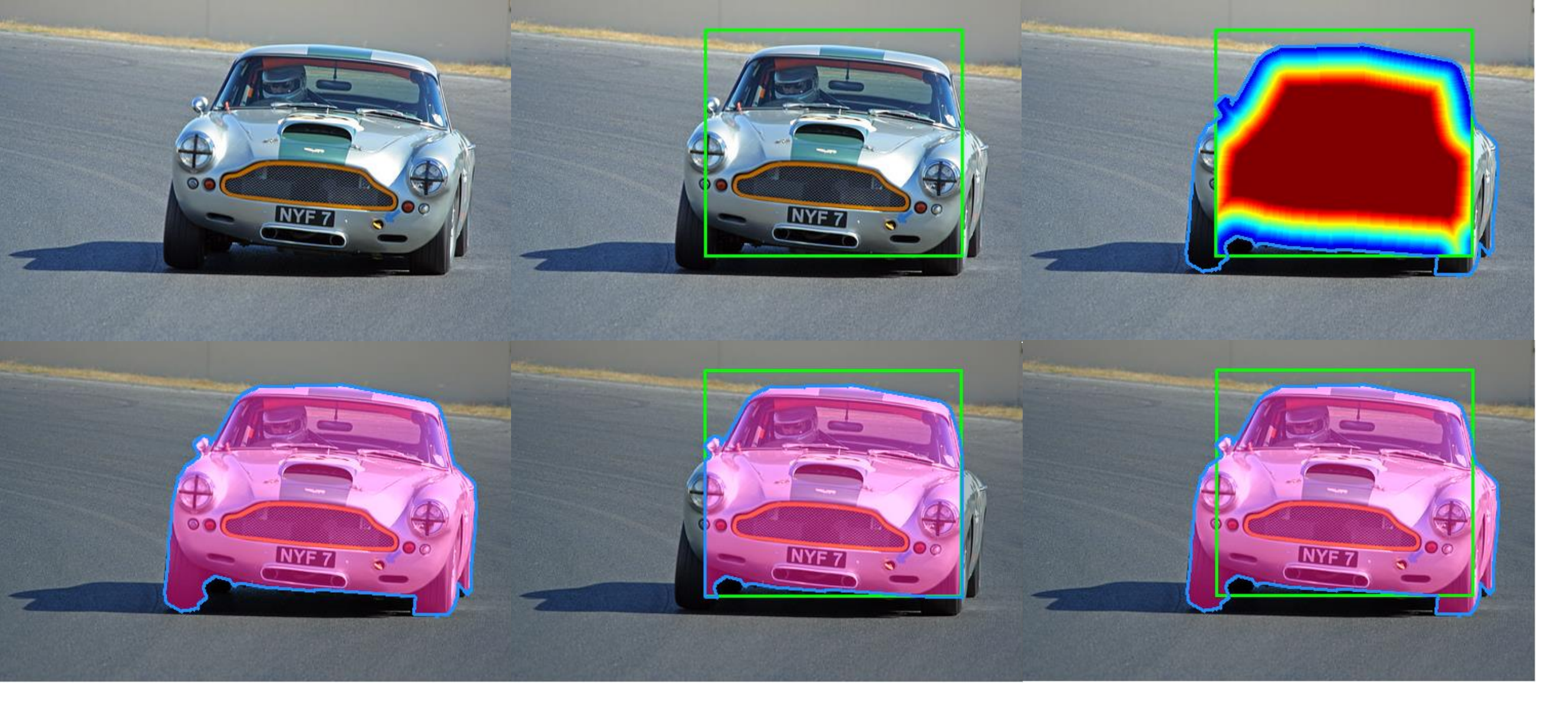}
	\caption{{\bf Traditional instance segmentation vs our boundary based representation.} {\bf Left:} Original image and ground-truth segmentation. {\bf Middle:} Given a bounding box, traditional methods directly predict a binary mask, whose extent is therefore limited to that of the box and thus suffers from box inaccuracies. {\bf Right:} We represent the object segment with a multi-valued map encoding the truncated minimum distance to the object boundary. This can be converted into a mask that goes beyond the bounding box, which makes our approach robust to box errors.}
	\vspace{-3mm}
	\label{fig:shape_repr}
\end{figure}

In this paper, we introduce a novel representation of object segments that is robust to errors in the bounding box proposals. To this end, we propose to model the shape of an object with a dense multi-valued map encoding, for every pixel in a box, its (truncated) minimum distance to the object boundary, or the fact that the pixel is outside the object. Object segmentation can then be achieved by converting this multi-valued map into a binary mask via the inverse distance transform~\cite{borgefors1986distance,kimmel1996sub}. In contrast to existing methods discussed above, and as illustrated in Fig.~\ref{fig:shape_repr}, the resulting mask is \emph{not} restricted to lie inside the bounding box;  even when the box covers only part of the object, the distances to the boundary in our representation may correspond to an object segment that goes beyond the box's spatial extent. 

To exploit our new object representation, we design an object mask network (OMN) that, for each box proposal, first predicts the corresponding pixel-wise multi-valued map, and then decodes it into the final binary mask, potentially going beyond the box itself. In particular, we discretize the truncated distances and encode them using a binary vector. 
This translates the prediction of the multi-valued map to a pixel-wise labeling task, for which deep networks are highly effective, and facilitates decoding the map into a mask. The first module of our network then produces multiple probability maps, each of which indicates the activation of one particular bit in this vector. We then pass these probability maps into a new residual-deconvolution network module that generates the final binary mask. Thanks to the deconvolution layers, our output is not restricted to lie inside the box, and our OMN is fully differentiable.

To tackle instance-level semantic segmentation, we integrate our OMN into the Multitask Network Cascade framework of~\cite{Jifeng15cocoseg}, by replacing the original binary mask prediction module. 
As our OMN is fully differentiable, we can learn the resulting instance-level semantic segmentation network in an end-to-end manner. Altogether, this yields a \textit{boundary-aware} instance segmentation (BAIS) network that is robust to noisy object proposals.

We demonstrate the effectiveness of our approach on PASCAL VOC 2012~\cite{Everingham12} and the challenging Cityscapes~\cite{Cordts2016Cityscapes} dataset. Our BAIS framework outperforms all the state-of-the-art methods on both datasets, by a considerable margin in the regime of high IOUs. Furthermore, an evaluation of our OMN on the task of object proposal generation on the PASCAL VOC 2012 dataset reveals that it achieves performance comparable to or even better than state-of-the-art methods, such as DeepMask~\cite{DeepMask} and SharpMask~\cite{SharpMask}.

\section{Related Work}

Over the years, much progress has been made on the task of category-level semantic segmentation, particularly since the advent of Deep Convolutional Neural Networks (CNNs)~\cite{farabet2013learning,FCN,chen2014semantic}. Categorical labeling, however, fails to provide detailed annotations of individual objects, from which many applications could benefit. By contrast, instance-level semantic segmentation produces information about the identity, location, shape and class label of each individual object.

To simplify this challenging task, most existing methods first rely on detecting individual objects, for which a detailed segmentation is then produced. The early instances of this approach~\cite{TigheNL14,he2014exemplar} typically used pre-trained class-specific object detectors. More recently, however, many methods have proposed to exploit generic object proposals~\cite{Arbelaez_2014_CVPR,fasterRCNN}, and postpone the classification problem to later stages. In this context,~\cite{HariharanAGM14} makes use of Fast-RCNN boxes~\cite{Girshick15} and builds a multi-stage pipeline to extract features, classify and segment the object. This framework was improved by the development of Hypercolumn features~\cite{HariharanAGM15} and the use of a fully convolutional network (FCN) to encode category-specific shape priors~\cite{LiHM15}. In~\cite{Jifeng15cocoseg}, the Region Proposal Network of~\cite{fasterRCNN} was integrated into a multi-task network cascade (MNC) for instance semantic segmentation. Ultimately, all these methods suffer from the fact that they predict a binary mask within the bounding box proposals, which are typically inaccurate. By contrast, here, we introduce a boundary-aware OMN that lets us predict instance segmentations that go beyond the box's spatial extent. We show that integrating this OMN in the MNC framework outperforms the state-of-the-art instance-level semantic segmentation techniques.

Other methods have nonetheless proposed to bypass the object proposal step for instance-level segmentation. For example, the Proposal-free Network (PFN) of~\cite{LiangArxiv15} predicts the number of instances and, at each pixel, a semantic label and the location of its enclosing bounding box. The results of this approach, however, strongly depend on the accuracy of the predicted number of instances. By contrast,~\cite{ZhangSFU15} proposed to identify the individual instances based on their depth ordering. This was further extended in~\cite{Ziyu15} via a deep densely connected Markov Random Field. It is unclear, however, how this approach handles the case where multiple instances are at roughly identical depths. To overcome this, the recent work of~\cite{UhrigCFB16} uses an FCN to jointly predict depth, semantics and an instance-based direction encoding. This information is then used to generate instances via a template matching procedure. Unfortunately, this process involves a series of independent modules, which cannot be optimized jointly, thus yielding a potentially suboptimal solution. Finally, in~\cite{Romera-ParedesT15}, a recurrent neural network was proposed to segment an image instance-by-instance. This approach, however, essentially assumes that all the instances observed in the image belong to the same class.

Beyond instance-level semantic segmentation, many methods have been proposed to generate class-agnostic region proposals~\cite{Arbelaez_2014_CVPR,Uijlings13,DBLP:conf/eccv/KrahenbuhlK14}. The most recent such approaches rely on deep architectures~\cite{DeepMask,SharpMask}. In particular, the method of~\cite{Jifeng16cocoseg}, in which an FCN computes a small set of instance-sensitive score maps that are assembled into object segment proposals, was shown to effectively improve instance-level semantic segmentation when incorporated in the MNC framework. Our experiments demonstrate that our OMN produces segments of a quality comparable to or even higher than these state-of-the-art methods. Furthermore, by integrating it in a complete instance-level semantic segmentation network, we also outperform the state-of-the-art on this task.


\begin{figure*}[t]
	\centering
	\begin{tabular}{lr}
	\includegraphics[trim = 0 0 680 0, clip, width=0.2\linewidth]{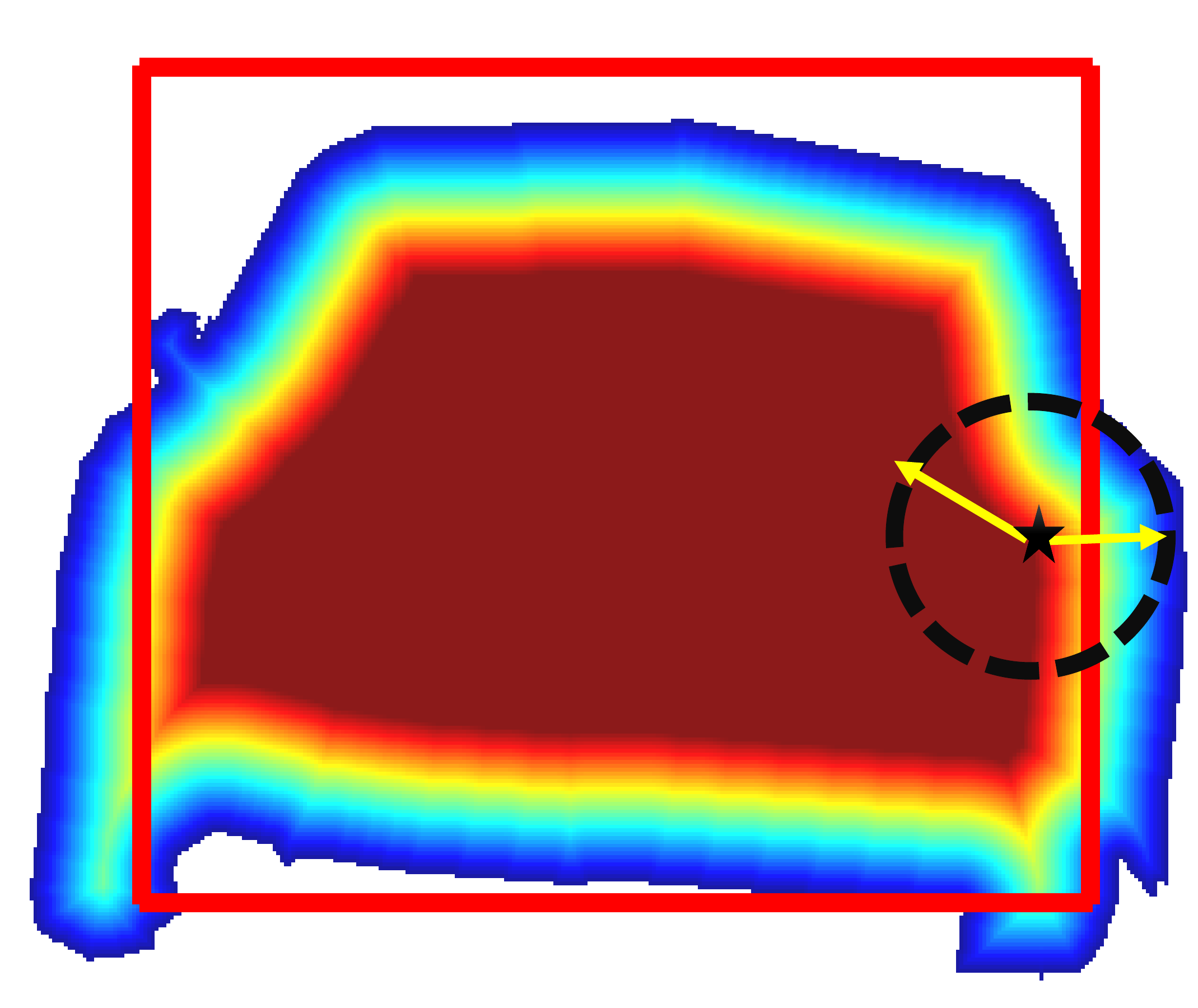}&	
	\includegraphics[trim = 0 150 50 0, clip, width=0.65\linewidth]{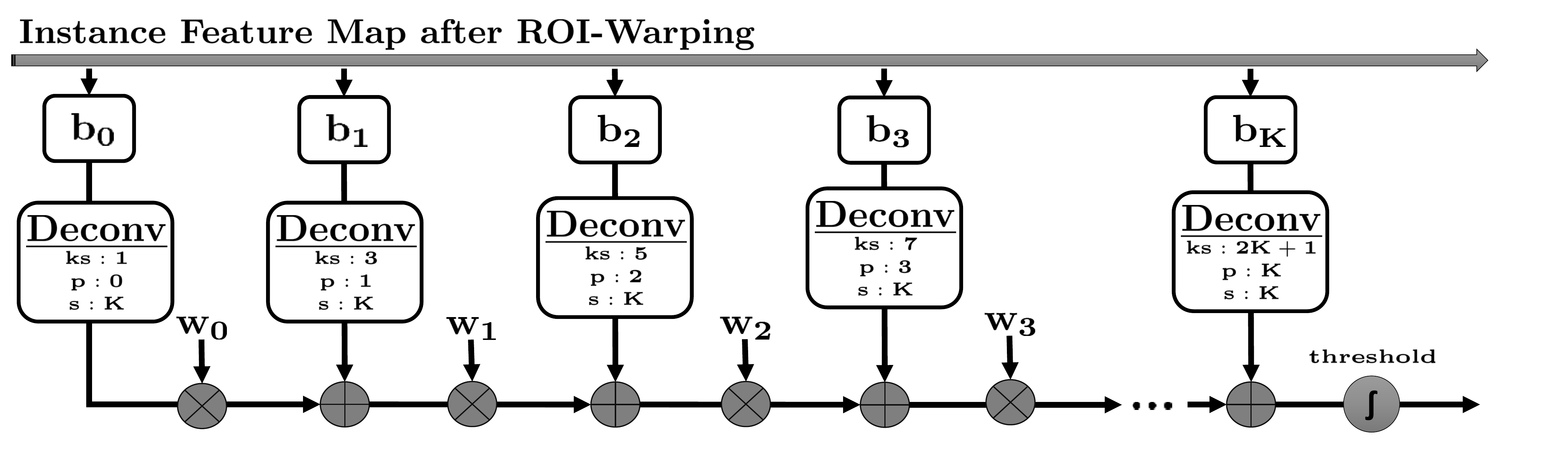}
	\end{tabular}
	\caption{{\bf Left:} Truncated distance transform. {\bf Right:} Our deconvolution-based shape-decoding network. Each deconvolution has a specific kernel size ($ks$), padding ($p$) and stride ($s$). Here, $K$ represents the number of binary maps.
	}\label{fig:ohe_module}
	\vspace{-4mm}	
\end{figure*}

\section{Boundary-aware Segment Prediction}
\label{sec:mask}
Our goal is to design an instance-level semantic segmentation method that is robust to the misalignment of initial bounding box proposals. To this end, we first introduce a novel object mask representation capable of capturing the overall shape or exact boundaries of an object. 
This representation, based on the distance transform, allows us to infer the complete shape of an object segment even when only partial information is available. We then construct a deep network that, given an input image, uses this representation to generate generic object segments that can go beyond the boundaries of initial bounding boxes. 

Below, we first describe our object mask representation and object mask network (OMN). In Section~\ref{sec:learn}, we show how our network can be integrated in a Multistage Network Cascade~\cite{Jifeng15cocoseg} to learn an instance-level semantic segmentation network in an end-to-end manner. 

\subsection{Boundary-aware Mask Representation}

Given a window depicting a potentially partially-observed object, obtained from an image and a bounding box, we aim to produce a mask of the entire object. To this end, instead of directly inferring a binary mask, which would only represent the visible portion of the object, we propose to construct a pixel-wise, multi-valued map encoding the boundaries of the complete object by relying on the concept of distance transform~\cite{borgefors1986distance}. In other words, the value at each pixel in our map represents either the distance to the nearest object boundary if the pixel is inside the object, or the fact that the pixel belongs to the background.

With varying window sizes and object shapes, the distance transform can produce a large range of different values, which would lead to a less invariant shape representation and complicate the training of our OMN in Section~\ref{sec:omn}. Therefore, we normalize the windows to a common size and truncate the distance transform to obtain a restricted range of values. 
Specifically, let $Q$ denote the set of pixels on the object boundary and outside the object. For every pixel $p$ in the normalized window, we compute a truncated distance $D(p)$ to $Q$ as
\begin{equation}
D(p) = \min\left(\min_{\forall q \in Q} \left\lceil{d(p, q)}\right \rceil, R\right)\;,
\label{equ:quantized_dt}
\end{equation}
where $d(p,q)$ is the spatial, Euclidean distance between pixel $p$ and $q$, $\lceil x \rceil$ returns the integer nearest to but larger than $x$, and $R$ is the truncation threshold, i.e., the largest distance we want to represent. We then directly use $D$ as our dense object representation. Fig.~\ref{fig:ohe_module} (Left) illustrates such a dense map for one object. 

As an object representation, the pixel-wise map described above as several advantages over a binary mask that specifies the presence or absence of an object of interest at each pixel. 
First, the value at a pixel gives us information about the location of the object boundary, even if the pixel belongs to the interior of the object. As such, our representation is robust to partial occlusions arising from inaccurate  bounding boxes. Second, since we have a distance value for every pixel, this representation is redundant, and thus robust to some degree of noise in the pixel-wise map. Importantly, predicting such a representation can be formulated as a pixel-wise labeling task, for which deep networks have proven highly effective.

To further facilitate this labeling task, we quantize the values in the pixel-wise map into $K$ uniform bins. In other words, we encode the truncated distance for pixel $p$ using a $K$-dimensional binary vector $b(p)$ as 
\begin{equation}
D(p) = \sum\limits_{n=1}^{K} r_n \cdot b_{n}(p), \quad \sum_{n=1}^K b_n(p) = 1\;,
\label{equ:quantized_dt_bits}
\end{equation}
where $r_n$ is the distance value corresponding to the $n$-th bin. By this one-hot encoding, we have now converted the multi-value pixel-wise map into a set of $K$ binary pixel-wise maps. This allows us to translate the problem of predicting the dense map to a set of pixel-wise binary classification tasks, which are commonly, and often successfully, carried out by deep networks. 

Given the dense pixel-wise map of an object segment (or truly $K$ binary maps), we can recover the complete object mask approximately by applying an inverse distance transform. Specifically, we construct the object mask by associating each pixel with a binary disk of radius $D(p)$. 
We then compute the object mask $M$ by taking the union of all the disks. Let $T(p,r)$ denote the disk of radius $r$ at pixel $p$. The object mask can then be expressed as
\begin{align}
M &= \bigcup\limits_{p} T(p,D(p)) = \bigcup_p T(p,\sum_{n=1}^K r_n\cdot b_n(p))\nonumber\\
&= \bigcup_{n=1}^K\bigcup\limits_{p} T(p,r_n\cdot b_n(p))=\bigcup_{n=1}^KT(\cdot,r_n)\ast B_n\;,
\label{equ:decode}
\end{align}
where $\ast$ denotes the convolution operator, and $B_n$ is the binary pixel-wise map for the $n$-th bin. Note that we make use of the property of the one-hot encoding in the derivation. Interestingly, the resulting operation consists of a series of convolutions, which will again become convenient when working with deep networks. 

The rightmost column in Fig.~\ref{fig:shape_repr} illustrates the behavior of our representation. In the top image, the value at each pixel represents the truncated distance to the instance boundary inside the bounding box. Although it does not cover the entire object, converting this dense map into a binary mask, yields the complete instance mask shown at the bottom.

\subsection{Object Mask Network}
\label{sec:omn}

We now turn to the problem of exploiting our boundary-aware representation to produce a mask for every object instance in an input image. To this end, we design a deep neural network that predicts $K$ boundary-aware dense binary maps for every box in a set of bounding box proposals and decodes them into a full object mask via Eq.~\ref{equ:decode}. In practice, we use the Region Proposal Network (RPN)~\cite{fasterRCNN} to generate the initial bounding box proposals. For each one of them, we perform a Region-of-Interest (RoI) warping of its features and pass the result to our network. This network consists of two modules described below. 

Given the RoI warped features of one bounding box as input, the first module in our network predicts the $K$ binary masks encoding our (approximate) truncated distance transform. Specifically, for the $n$-th binary mask , we use a fully connected layer with a sigmoid activation function to predict a pixel-wise probability map that approximates $B_n$. 

Given the $K$ probability maps, we design a new residual deconvolution network module to decode them into a binary object mask. Our network structure is based on the observation that the morphology operator in Eq.~\ref{equ:decode} can be implemented as a series of deconvolutions with fixed weights but different kernel and padding sizes, as illustrated in Fig.~\ref{fig:ohe_module}~(Right). We then approximate the union operator with a series of weighted summation layers followed by a sigmoid activation function. The weights in the summation layers are learned during training.
To accommodate for the different sizes of the deconvolution filters, we upsample the output of the deconvolution corresponding to a smaller value of $r_n$ in the network before each weighted summation. We use a fixed stride value of $K$ for this purpose. 

Our OMN is fully differentiable, and the output of the decoding module can be directly compared to the ground truth at a high resolution using a cross-entropy loss. This allows us to train our OMN in an end-to-end fashion, including the initial RPN, or, as discussed in Section~\ref{sec:learn}, to integrate it with a classification module to perform instance-level semantic segmentation. 
\section{Learning Instance Segmentation} 
\label{sec:learn}
We now introduce our approach to tackling instance-level semantic segmentation with our OMN. To this end, we construct a Boundary-Aware Instance Segmentation (BAIS) network by integrating our object mask network into a Multistage Network Cascade (MNC)~\cite{Jifeng15cocoseg}. Since our OMN module is differentiable, we can train the entire instance segmentation network in an end-to-end manner. Below, we first describe the overall network architecture, and then discuss our end-to-end training procedure and inference at test time.

\begin{figure*}[t]
\centering
\begin{tabular}{lr}
\includegraphics[trim = 0 350 550 0, clip, width=0.65\linewidth]{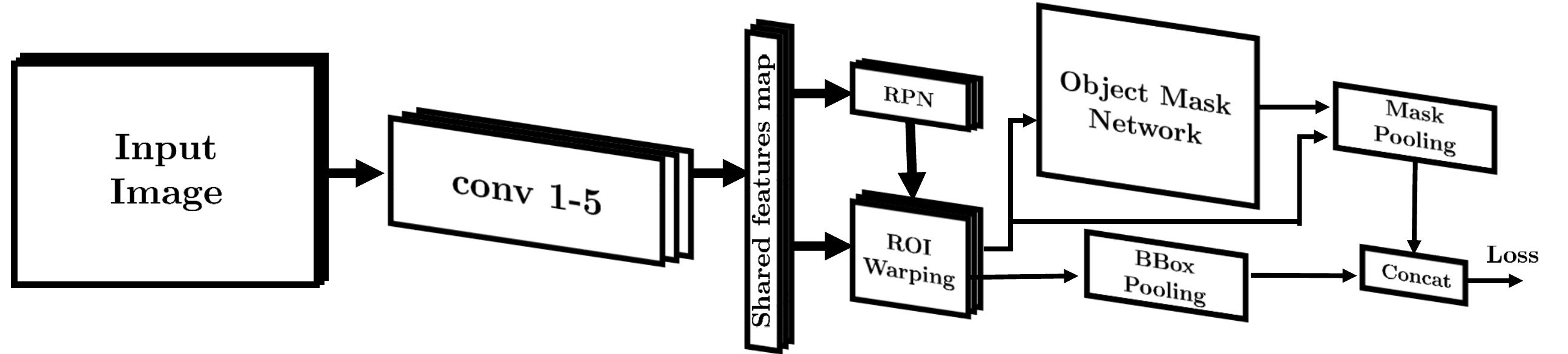}
\includegraphics[trim = 0 320 870 0, clip, width=0.33\linewidth]{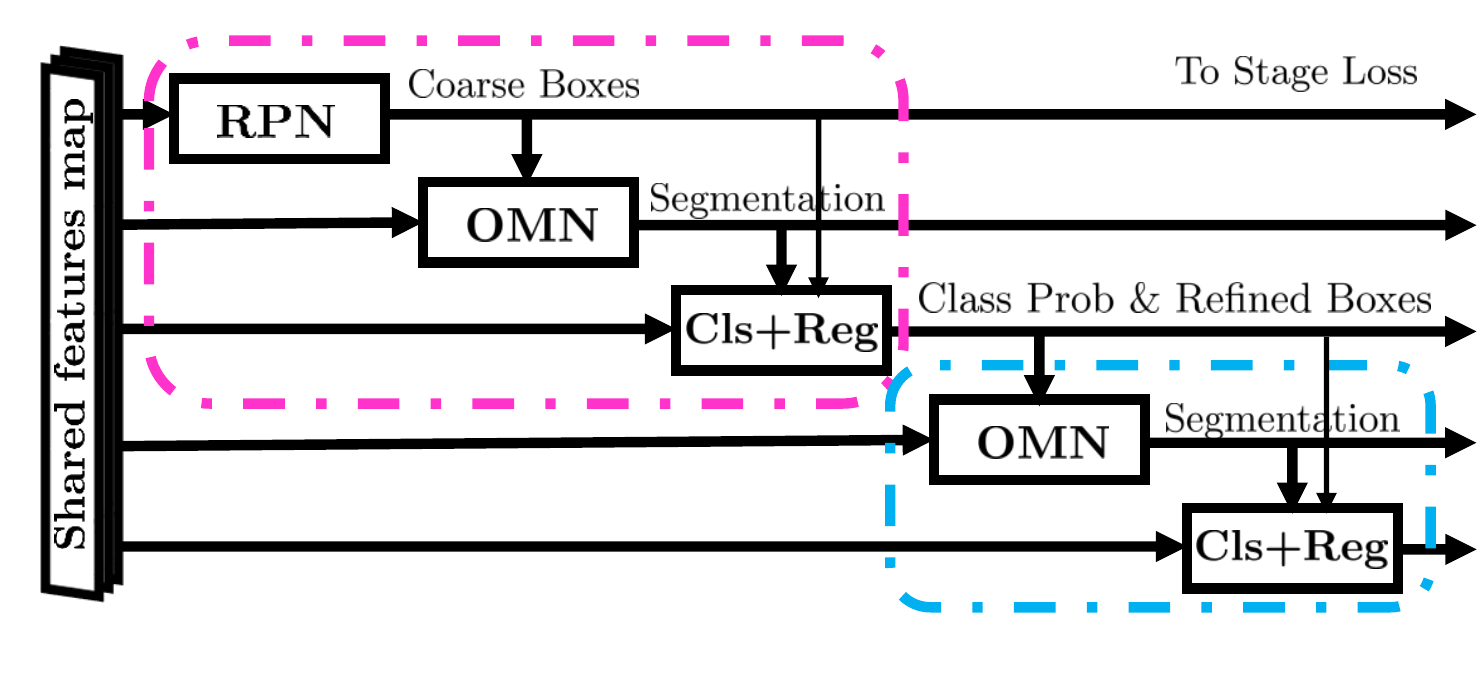}
\end{tabular}
\caption{\X{Left:} Detailed architecture of our boundary-aware instance segmentation network. An input image first goes through a series of convolutional layers, followed by an RPN to generate bounding box proposals. After RoI warping, each proposal passes through our OMN to obtained a binary mask that can go beyond the box's spatial extent. Mask features are then extracted and used in conjunction with bounding-box features for classification purpose. During training, our model makes use of a multi-task loss encoding bounding box, segmentation and classification errors. \X{Right:} 5-stage BAIS network. The first three stages correspond to the model on the left. The five-stage model then concatenates an additional OMN  and classification module to these three stages. The second OMN takes as input the classification score and refined box from the previous stage, and outputs a new segmentation with a new score obtained via the second classification module. The weights of the OMN and classification modules in both stages are shared. 
}\vspace{-4mm}
\label{fig:network}
\end{figure*}

\subsection{BAIS Network}

Our boundary-aware instance segmentation network follows a structure similar to that of the MNC. Specifically, our segmentation network consists of three sub-networks, corresponding to the tasks of bounding box proposal generation, object mask prediction and object classification. The first module consists of a deep CNN (in practice, the VGG16~\cite{simonyan2014very} architecture) to extract a feature representation from an input image, followed by an RPN~\cite{fasterRCNN}, which generates a set of bounding box proposals. After RoI warping, we pass each proposal through our OMN to produce a segment mask. 
Finally, as in the original MNC network, mask features are computed by using the predicted mask in a feature masking layer and concatenated with bounding box features. The resulting representation is then fed into the third sub-network, which consists of a single fully-connected layer for classification and bounding-box regression. 
The overall architecture of our BAIS network is illustrated in Fig.~\ref{fig:network}. 

\vspace{-3mm}
\paragraph{Multi-stage Boundary-aware Segmentation Network.} 
Following the strategy of~\cite{Jifeng15cocoseg}, we extend the BAIS network described above, which can be thought of as a 3-stage cascade, to a 5-stage cascade. 
The idea, here, is to refine the initial set of bounding box proposals, and thus the predicted segments, based on the output of our OMN. As illustrated in Fig.~\ref{fig:network} (Right), the first three stages consist of the model described above, that is the VGG16 convolutional layers, RPN, OMN, classification module and bounding-box prediction. 
We then make use of the prediction offset generated by the bounding-box regression part of the third stage to refine the initial boxes. 
These new boxes act as input, via RoI warping, to the fourth-stage, which corresponds to a second OMN. Its output is then used in the last stage in conjunction with the refined boxes for classification purpose. In this 5-stage cascade, the weights of the two OMN and of the two classification modules are shared. 

\subsection{Network Learning and Inference}
Our BAIS network is fully differentiable, and we therefore train it in an end-to-end manner. To this end, we use a multi-task loss function to account for bounding box, object mask and classification errors. Specifically, we use the softmax loss for the RPN and for classification, and the binary cross-entropy loss for the OMN. 
In our five-stage cascade, the bounding box and mask losses are computed after the third and fifth stages, and we use the smooth $L_1$ loss for bounding-box regression. 

We minimize the resulting multi-task, multi-stage loss over all parameters jointly using stochastic gradient descent (SGD). Following~\cite{Jifeng15cocoseg,Jifeng16cocoseg,Girshick15}, we rely on min-batches of 8 images. 
As in~\cite{Jifeng15cocoseg,fasterRCNN,Girshick15}, we resize the images such that the shorter side has 600 pixels. The VGG16 network in our first module was pre-trained on ImageNet. The other weights are initialized randomly from a zero-mean Gaussian distribution with std 0.01. We then train our model for 20k iterations with a learning rate of 0.001, and 5k iterations with a reduced learning rate of 0.0001. 

The first module in our network first generates $\sim$12k bounding boxes, which are pruned via non-maximum suppression (NMS). As in~\cite{Jifeng15cocoseg}, we use an NMS threshold of 0.7, and finally keep the top 300 bounding box proposals. 
In our OMN, we use $K = 5$ probability maps to encode the (approximate) truncated distance transform. After decoding these maps via Eq.~\ref{equ:decode}, we make use of a threshold of 0.4 to obtain a binary mask. This mask is then used to pool the features, and we finally obtain the semantic label via the classification module.

At test time, our BAIS network takes an input image and first computes the convolutional feature maps. The RPN module then generates 300 bounding box proposals and our OMN module predicts the corresponding object masks. These masks are categorized according to the class scores and a class-specific non-maximum suppression is applied with an IoU threshold of 0.5. Finally, we apply the in-mask voting scheme of~\cite{Jifeng15cocoseg} to each category independently to further refine the instance segmentations. \note{This is a bit obscure, but maybe it's ok.}

\section{Experiments}
In this section, we demonstrate the effectiveness of our method on instance-level semantic segmentation and segment proposal generation. We first discuss the former, which is the main focus of this work, and then turn to the latter. In both cases, we compare our approach to the state-of-the-art methods in each task.

\setcounter{table}{2}  
\begin{table*}[!b]
\begin{center}
\resizebox{0.8\linewidth}{!}{%
\begin{tabular}{r|cccc}
\hline
Cityscapes (test) & AP & AP (50\%) & AP (100m) & AP (50m) \Tstrut \\ 
\hline
Instance-level Segmentation of Vehicles by Deep Contours \cite{JanvandenBrand2016} & 2.3 & 3.7 & 3.9 & 4.9 \Tstrut\\
R-CNN + MCG convex hull \cite{Cordts2016Cityscapes} & 4.6 & 12.9 & 7.7 & 10.3 \\
Pixel-level Encoding for Instance Segmentation \cite{UhrigCFB16} & 8.9 & 21.1 & 15.3 & 16.7 \\
RecAttend \cite{RecAttend}& 9.5 & 18.9 & 16.8 & 20.9 \\
InstanceCut \cite{kirillov-2016-arxiv}& 13.0 & 27.9 & 22.1 & 26.1 \\
DWT \cite{BaiU16}& 15.6 & 30.0 & 26.2 & 31.8 \\
\hline
BAIS - full (ours) & \X{17.4} & \X{36.7} & \X{29.3} & \X{34.0} \Tstrut \Bstrut \\
\hline
\end{tabular}}
\end{center} \vspace{-0.2cm}
\caption{{\bf Instance-level semantic segmentation on Cityscapes.} We compare our method with the state-of-the-art baselines on the Cityscapes test set. These results were obtained from the online evaluation server.} 
\label{res:cs_instance_all}
\end{table*}

\vspace{-0.4cm}
\paragraph{Datasets and setup.}
To evaluate our approach, we make use of two challenging, standard datasets with multiple instances from a variety of object classes, \ie, Pascal VOC 2012 and Cityscapes.

The Pascal VOC 2012 dataset~\cite{Everingham12} comprises 20 object classes with instance-level ground-truth annotations for 5623 training images and 5732 validation images. We used the instance segmentations of~\cite{BharathICCV2011} for training and validation. We used all the training images to learn our model, but, following the protocols used in~\cite{HariharanAGM14, HariharanAGM15, DaiH015, Jifeng15cocoseg, Jifeng16cocoseg}, used only the validation dataset for evaluation. Following standard practice, we report the mean Average Precision (mAP) using IoU thresholds of 0.5 and 0.7 for instance semantic segmentation, and the Average Recall (AR) for different number and sizes of boxes for segment proposal generation.

The Cityscapes dataset~\cite{Cordts2016Cityscapes} consists of 9 object categories for instance-level semantic labeling. This dataset is very challenging since each image can contain a much larger number of instances of each class than in Pascal VOC, most of which are very small. It comprises 2975 training images from 18 cities, 500 validation images from 3 cities and 1525 test images from 6 cities. We only used the training dataset for training, and the test dataset to evaluate our method's performance on the online test-server. Following the Cityscapes dataset guidelines, we computed the average precision (AP) for each class by averaging it across a range of overlap thresholds. We report the mean average precision (mAP) using an IoU threshold of 0.5, as well as mAP100m and mAP50m, where the evaluation is restricted to objects within 100 meters and 50 meters, respectively.

\note{There is a serious issue with the placement of the tables. The should appear in the correct order and be reasonably spread over the experiments section. That said, I don't know how to fix it. Please try.}

\subsection{Instance-level Semantic Segmentation}
We first present our results on the task of instance-level semantic segmentation, which is the main focus of this paper. We report results on the two datasets discussed above. In both cases, we restricted the number of proposals to 300. For our 5-stage model, this means 300 after the first RPN and 300 after bounding-box refinement.

\subsubsection{Results on VOC 2012}
Let us first compare the results of our Boundary-aware Instance Segmentation (BAIS) network with the state-of-the-art approaches on Pascal VOC 2012. These baselines include the SDS framework of~\cite{HariharanAGM14}, the Hypercolumn representation of~\cite{HariharanAGM15}, the InstanceFCN method of~\cite{Jifeng16cocoseg} and the MNC framework of~\cite{Jifeng15cocoseg}. In addition to this, we also report the results obtained by a Python re-implementation of the method in \cite{Jifeng15cocoseg}, which we refer to as MCN-new. The results of this comparison are provided in Table~\ref{res:voc_2012_instance_all}. Note that our approach outperforms all the baselines, by a considerable margin in the case of a high IOU threshold. Note also that our approach is competitive in terms of runtime. Importantly, the comparison with BAIS-inside BBox, which restricts our masks to the spatial extent of the bounding boxes clearly evidences the importance of allowing the masks to go beyond the boxes' extent.

\setcounter{table}{0}  
\begin{table}[!t]
\begin{center}
\resizebox{\linewidth}{!}{
\begin{tabular}{r|ccc}
\hline
VOC 2012 (val) 									& mAP (0.5) & mAP (0.7) & time/img (s) 	\Tstrut \\
\hline
SDS \cite{HariharanAGM14}						& 49.7		& 25.3		& 48	\Tstrut \\
PFN \cite{LiangArxiv15}							& 58.7		& 42.5		& $\sim$ 1		\\
Hypercolumn \cite{HariharanAGM15}				& 60.0		& 40.4		& \textgreater 80	\\
InstanceFCN \cite{Jifeng16cocoseg} 				& 61.5 		& 43.0 		& 1.50	\\
MNC \cite{Jifeng15cocoseg} 						& 63.5 		& 41.5 		& 0.36	\\
MNC-new 	& 65.01 	& 46.23 	& 0.42	\Bstrut \\
\hline
BAIS - insideBBox (ours)						& 64.97 	& 44.58 	& 0.75	\Tstrut \\
BAIS - full (ours) 								& \X{65.69} & \X{48.30} & 0.78	\\
\hline
\end{tabular}}
\end{center} \vspace{-0.2cm}
\caption{{\bf Instance-level semantic segmentation on Pascal VOC 2012.} Comparison of our method with state-of-the-art baselines. The results of \cite{HariharanAGM14, HariharanAGM15} are reproduced from~\cite{Jifeng15cocoseg}.}
\label{res:voc_2012_instance_all}
\end{table}

\begin{table}[!t]
\begin{center}
\resizebox{\linewidth}{!}{
\begin{tabular}{r|cc|cc}
\hline
VOC 2012 (val) 									& Training 	& Testing 	& mAP (0.5) & mAP (0.7) \Tstrut \\ 
\hline
MNC \cite{Jifeng15cocoseg} 						& 3 stage 	& 5 stage 	& 62.6 		& - 		\Tstrut \\
MNC-new 	& 5 stage 	& 5 stage 	& 65.01 	& 46.23 	\Bstrut \\
\hline
BAIS - full (ours) 								& 3 stage 	& 5 stage 	& 65.51 	& 47.13 	\Tstrut \\
BAIS - full (ours) 								& 5 stage 	& 5 stage 	& \X{65.69} & \X{48.30} \Bstrut \\
\hline
\end{tabular}}
\end{center} \vspace{-0.2cm}
\caption{{\bf Influence of the number of stages during training.} Whether trained using 3 stages or 5, our approach outperforms both MNC baselines.} 
\label{res:voc_2012_instance_results_mnc}\vspace{-2mm}
\end{table}
\setcounter{table}{3}  

\begin{table*}[!t]
\begin{center}
\resizebox{0.7\linewidth}{!}{
\begin{tabular}{r|cccccccc|c}
\hline
Cityscapes (test) 		& person & rider & car & truck & bus & train & motorcycle & bicycle & AP (50m) 	\Tstrut \\ \hline
DWT 					& 27.0 & 20.4 & 57.0 & \X{39.6} & \X{51.3} & 37.9 & 12.8 & 8.1 & 31.8 \Tstrut	\\
BAIS - full (ours) 		& \X{31.5} & \X{23.4} & \X{63.1} & 32.2 & 50.5 & \X{40.4} & \X{16.5} & \X{14.6} & \X{34.0} \Bstrut \\ \hline
\hline
Cityscapes (test) 		& person & rider & car & truck & bus & train & motorcycle & bicycle & AP (100m) 	\Tstrut \\ \hline
DWT 					& 27.0 & 19.8 & 52.8 & \X{29.0} & 36.4 & 25.1 & 11.7 & 7.8 & 26.2 \Tstrut	\\
BAIS - full (ours) 		& \X{30.3} & \X{22.7} & \X{58.2} & 24.9 & \X{38.6} & \X{29.9} & \X{15.3} & \X{14.3} & \X{29.3} \Bstrut \\ \hline
\end{tabular}}
\end{center} \vspace{-0.2cm}
\caption{{\bf Detailed comparison with DTW: Top:} AP(50m), {\bf Bottom:} AP(100m). Note that our approach outperforms this baseline on all the classes except truck for the Cityscapes test dataset.}
\label{res:cs_instance_results_50m_100m}\vspace{-2mm}
\end{table*}

\begin{table*}[!t]
\begin{center}
\resizebox{0.75\linewidth}{!}{
\begin{tabular}{r|c|cccccccc|c}
\hline
Cityscapes (val) 								& IoU & person & rider & car & truck & bus & train & motorcycle & bicycle & mAP 	\Tstrut \\ \hline
MNC-new 	& 0.5 & 23.25 & 25.19 & 43.26 & 31.65 & 50.99 & 42.51 & 14.00 & 17.53 & 31.05 	\Tstrut \\
BAIS - full (ours) 								& 0.5 & 23.30 & 25.67 & 43.19 & 33.01 & 54.36 & 44.87 & 15.95 & 18.84 & \X{32.40} \Bstrut \\ \hline
MNC-new 	& 0.7 & 9.09 & 1.86 & 34.81 & 24.46 & 39.08 & 33.33 & 1.98 & 4.55 & 18.64 	\Tstrut \\
BAIS - full (ours) 								& 0.7 & 9.09 & 2.53 & 35.05 & 25.75 & 39.35 & 33.04 & 2.73 & 5.30 & \X{19.10} \Bstrut \\
\hline
\end{tabular}}
\end{center} \vspace{-0.2cm}
\caption{{\bf Comparison with MNC-new on the Cityscapes validation data.} Note that our approach outperforms this baseline, thus showing the importance of allowing the masks to go beyond the box proposals.} 
\label{res:cs_instance_results_mnc}\vspace{-2mm}
\end{table*}

Following the evaluation of MNC in~\cite{Jifeng15cocoseg}, we also study the influence of the number of stages in our model. We therefore learned different versions of our model using either our three-stage or five-stage cascade. At test time, because of parameter sharing across the stages, both versions are tested following a 5-stage procedure. 
The results of these different training strategies, for both MNC and our approach, are shown in Table~\ref{res:voc_2012_instance_results_mnc}. Note that, while our model trained with five-stages achieves the best results, our three-stage model still outperforms the two MNC baselines. 

A detailed comparison with MNC~\cite{Jifeng15cocoseg} including results for all the classes is provided in the supplementary material.

\vspace{-2mm}
\subsubsection{Results on Cityscapes}
We now turn to the Cityscapes dataset. In Table~\ref{res:cs_instance_all}, we first report the results obtained from the online evaluation server on the test data, which is not publicly available. Note that our approach outperforms all the baselines significantly on all the metrics. In Table~\ref{res:cs_instance_results_50m_100m}, we provide a detailed comparison of our approach and the best performing baseline (DWT) in terms of AP(100m) and AP(50m), respectively. Note that we outperform this method on most classes.

Additionally, we also compare our approach with MNC-new on the validation data. In this case, both models were trained using the training data only. For MCN, we used the same image size, RPN batch size, learning rate and number of iterations as for our model. Both models were trained using 5 stages. Table~\ref{res:cs_instance_results_mnc} shows that, again, our model outperforms this baseline, thus demonstrating the benefits of allowing the masks to go beyond the box proposals.

\begin{figure*}[!t]
\begin{center}
\includegraphics[trim=1.8cm 3.0cm 1.8cm 3.0cm, clip=true, width=.85\linewidth]{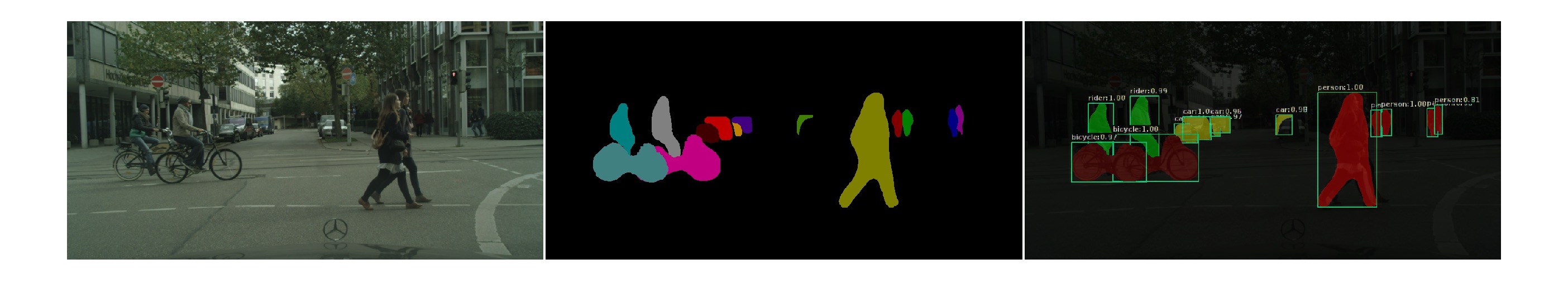} \\
\includegraphics[trim=1.8cm 3.0cm 1.8cm 3.0cm, clip=true, width=.85\linewidth]{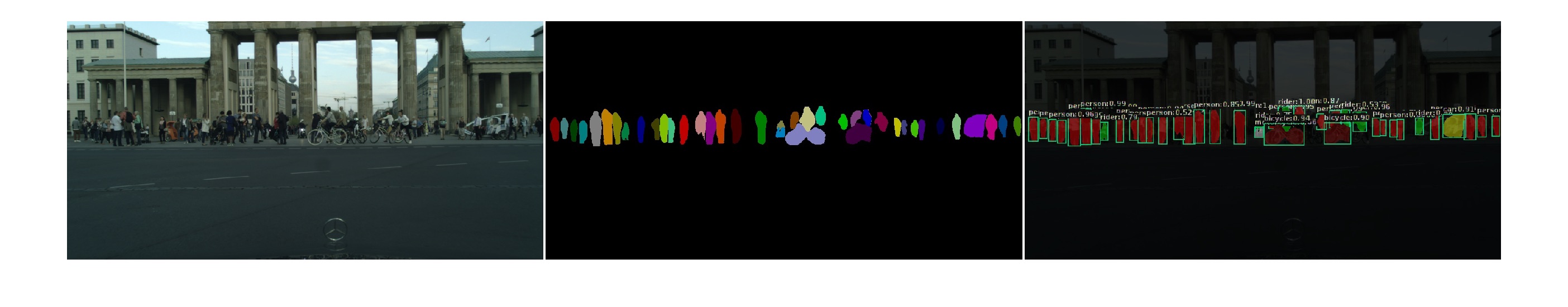} \\
\includegraphics[trim=1.8cm 3.0cm 1.8cm 3.0cm, clip=true, width=.85\linewidth]{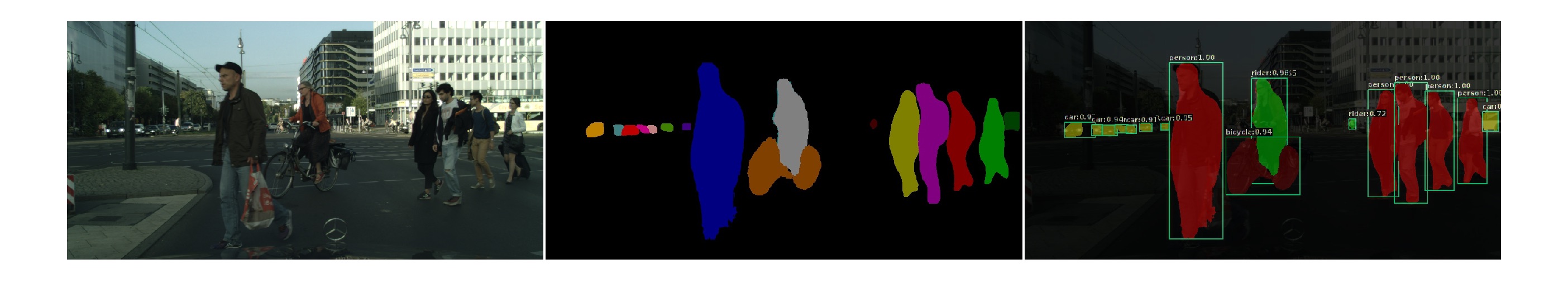} \\
\includegraphics[trim=1.8cm 3.0cm 1.8cm 3.0cm, clip=true, width=.85\linewidth]{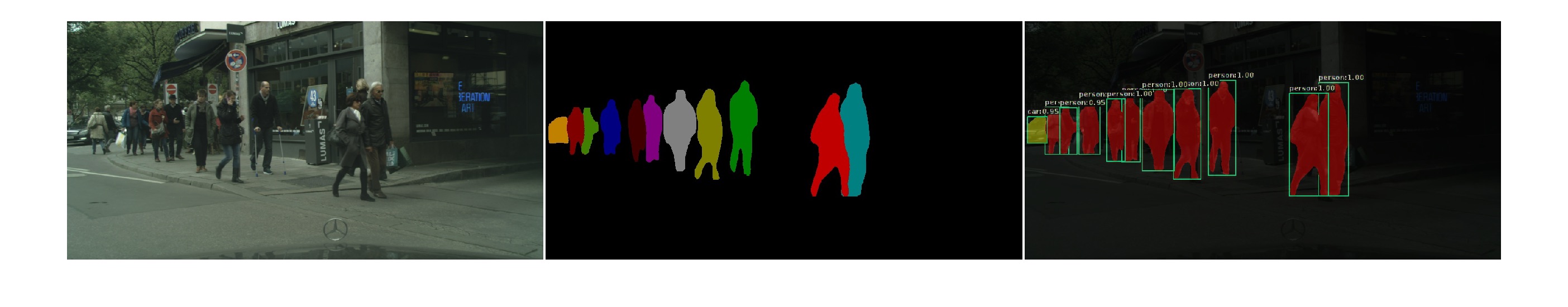} \\
\includegraphics[trim=1.8cm 3.0cm 1.8cm 3.0cm, clip=true, width=.85\linewidth]{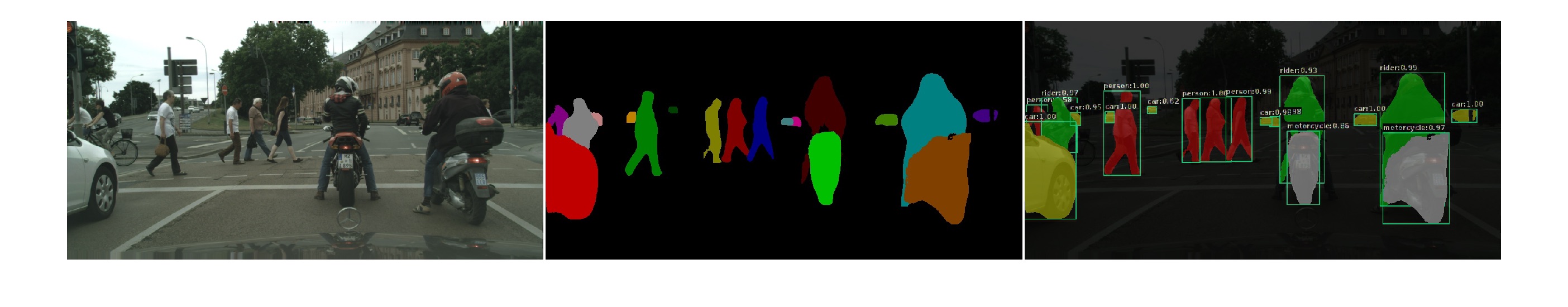} \\
\end{center} \vspace{-0.3cm}
\caption{\textbf{Qualitative results on Cityscapes.} From left to right, we show the input image, our instance level segmentations and the segmentations projected onto the image with class labels. Note that our segmentations are accurate despite the presence of many instances.} 
\label{fig:cs_test_qualitative}
\end{figure*}

\begin{figure*}[ht]
\begin{center}
\includegraphics[trim=1.8cm 3.0cm 1.8cm 3.0cm, clip=true, width=.85\linewidth]{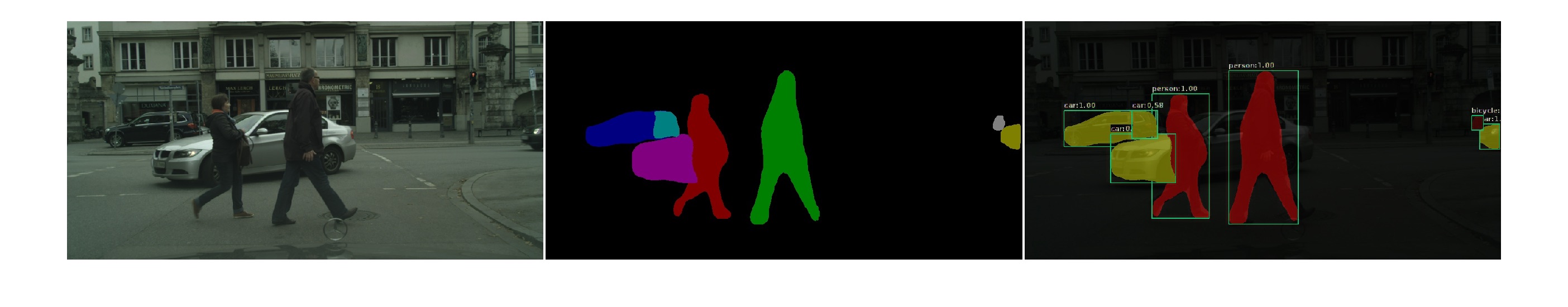} \\
{\includegraphics[trim=1.8cm 3.0cm 1.8cm 3.0cm, clip=true, width=.85\linewidth]{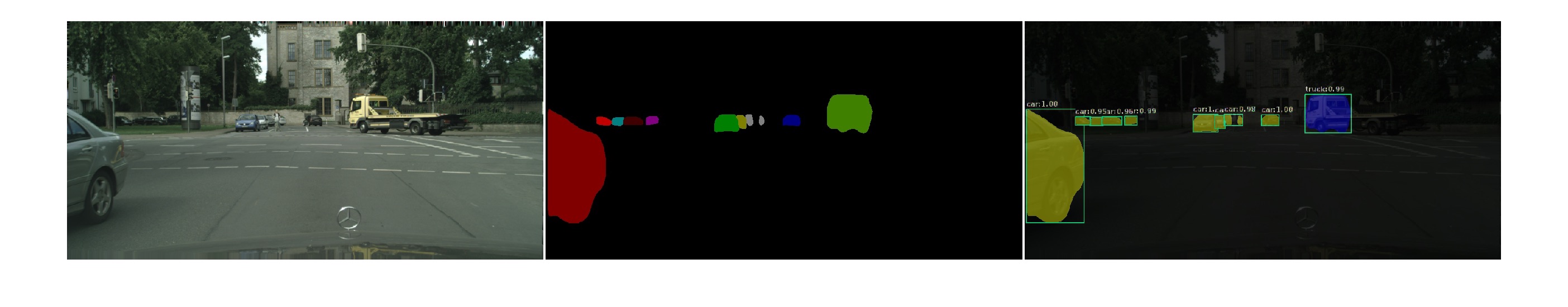}} \\
\end{center} \vspace{-0.3cm}
\caption{\textbf{Failure cases.} The typical failures of our approach correspond to cases where one instance is broken into multiple ones.}
\label{fig:cs_test_qualitative_fail}
\end{figure*}

In Fig.~\ref{fig:cs_test_qualitative}, we provide some qualitative results of our approach on Cityscapes. Note that we obtain detailed and accurate segmentations, even in the presence of many instances in the same image. Some failure cases are shown in Fig.~\ref{fig:cs_test_qualitative_fail}. These failures typically correspond to one instance being broken into several ones.

\subsection{Segment Proposal Generation}
As a second set of experiments, we evaluate the effectiveness of our object mask network (OMN) at generating high-quality segment proposals. To this end, we made use of the  5732 Pascal VOC 2012 validation images with ground-truth from~\cite{BharathICCV2011}, and compare our approach with the state-of-the-art segmentation proposal generation methods according to the criteria of~\cite{HariharanAGM14, DBLP:COCO14}. In particular, we report the results of MCG~\cite{Arbelaez_2014_CVPR}, Deep-Mask~\cite{DeepMask} and Sharp-Mask~\cite{SharpMask} using the publicly available pre-computed segmentation proposals. We also report the results of MNC by reproducing them from~\cite{Jifeng15cocoseg}, since these values were slightly better than those obtained from the publicly available segments. For our method, since the masks extend beyond the bounding box, the scores coming from the RPN, which correspond to the boxes, are ill-suited. We therefore learned a scoring function to re-rank our proposals. For the comparison to be fair, we also learned a similar scoring function for the MNC proposals. We refer to this baseline as MNC+score. 

\begin{figure*}[!b]
\begin{center} \vspace{-1cm}
\begin{subfigure}{\includegraphics[trim=0cm 0cm 0cm 0cm, clip=true, width=.24\linewidth]{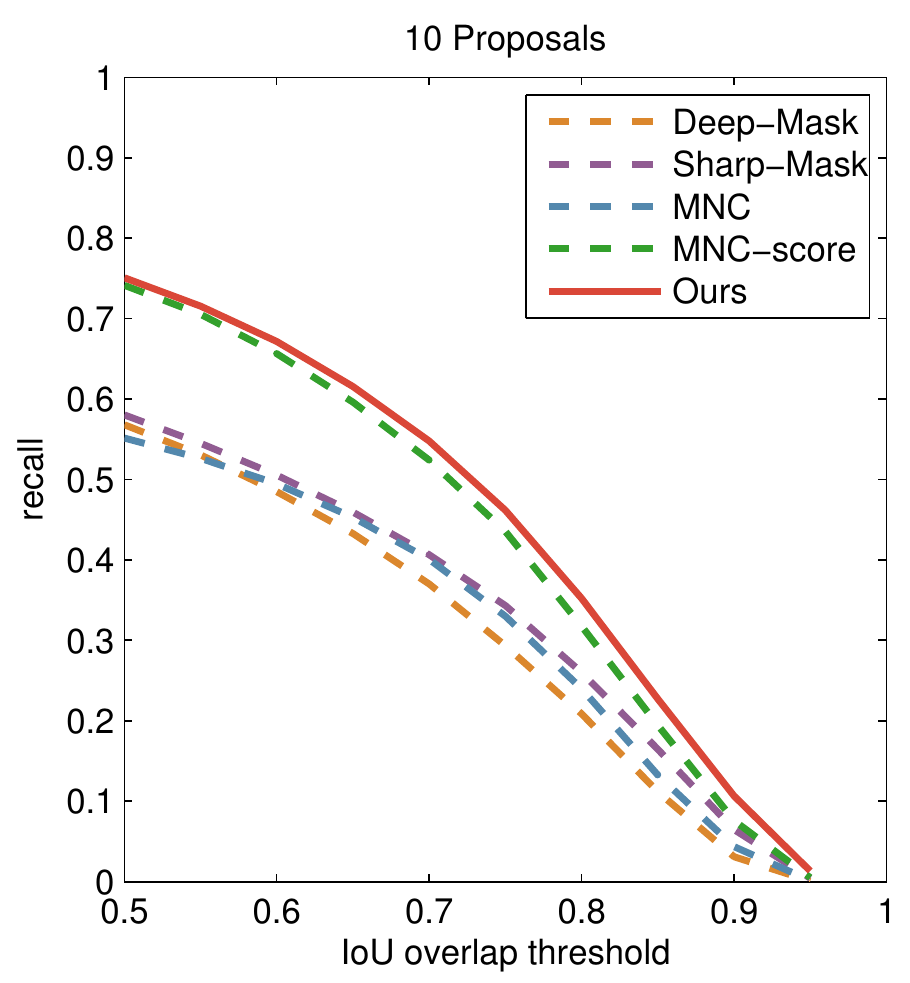}}\end{subfigure}
\begin{subfigure}{\includegraphics[trim=0cm 0cm 0cm 0cm, clip=true, width=.24\linewidth]{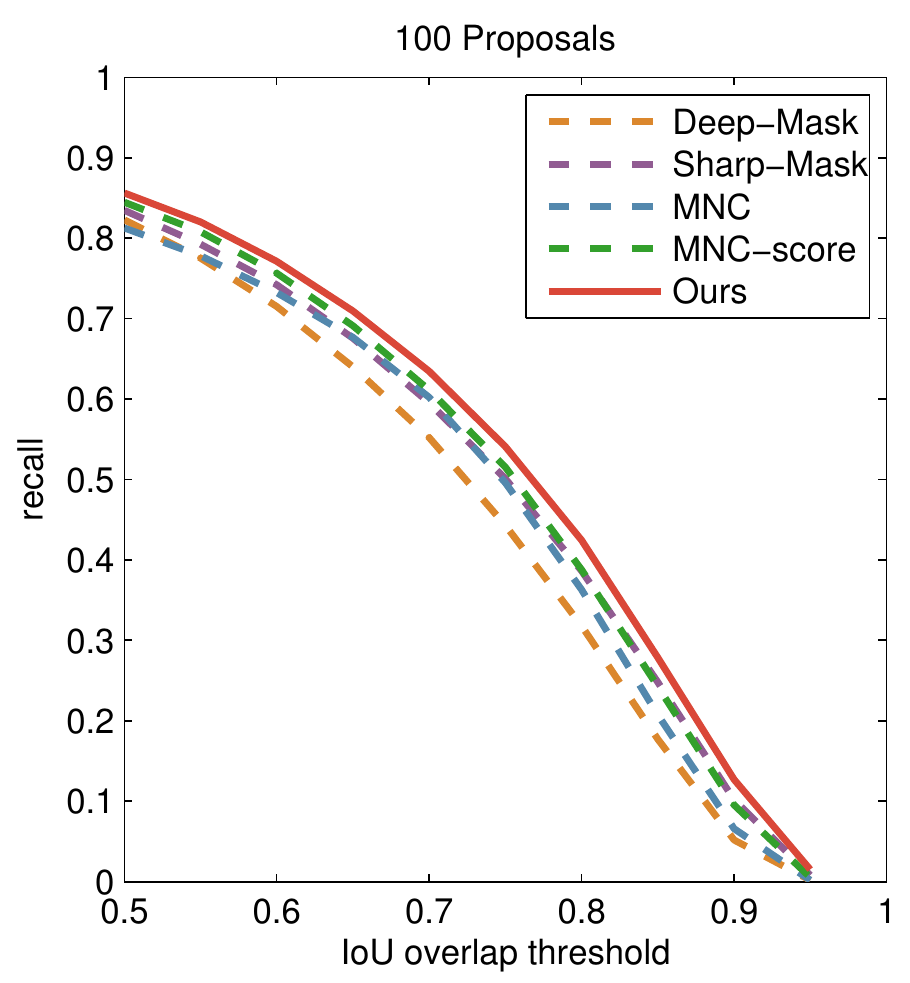}}\end{subfigure}
\begin{subfigure}{\includegraphics[trim=0cm 0cm 0cm 0cm, clip=true, width=.24\linewidth]{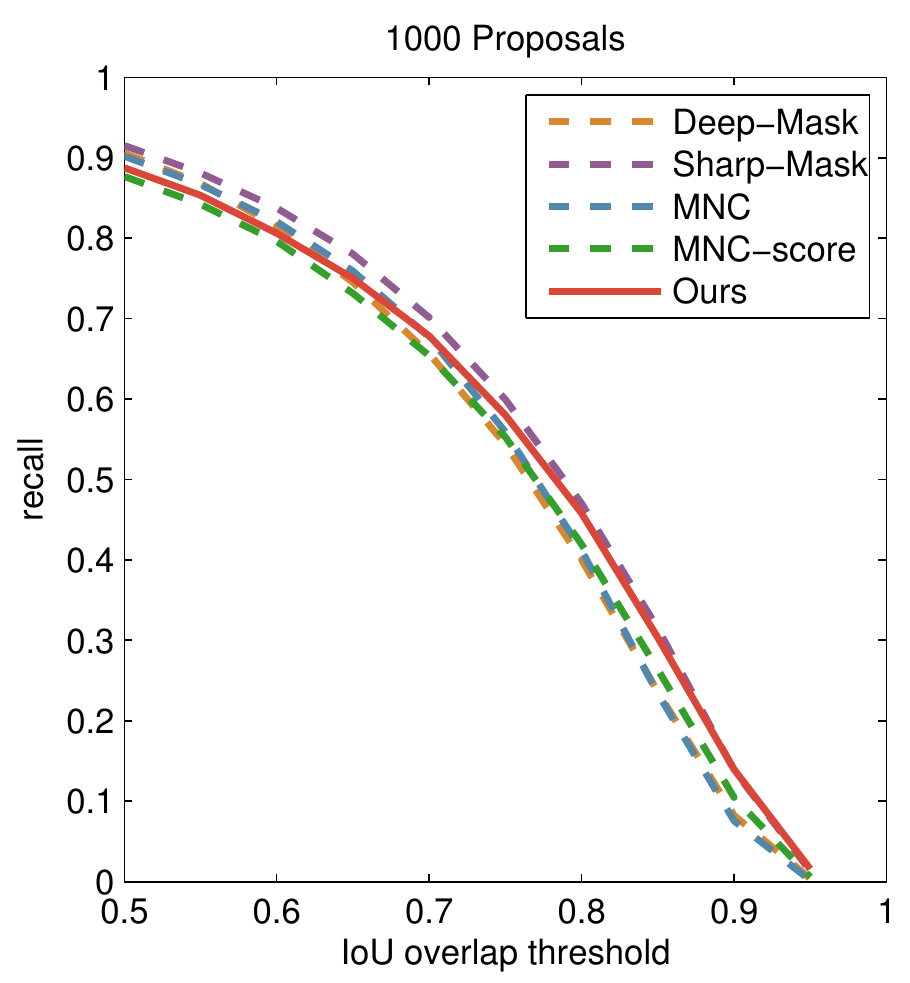}}\end{subfigure}
\end{center} \vspace{-0.2cm}
\caption{
\textbf{Recall v.s. IoU threshold on Pascal VOC 2012.} The curves were generated using the highest-scoring 10, 100 and 1000 segmentation proposals, respectively. In each plot, the solid line corresponds to our OMN results. Note that we outperform the baselines when using the top 10 and 100 proposals. For 1000, our approach still yields state-of-the-art results at high IoU thresholds.}
\label{fig:voc_2012_val_recall} 
\end{figure*}

The results of our comparison are provided in Table~\ref{tab:voc_2012_val_table}. Our approach yields state-of-the-art results when considering 10 or 100 proposals. For 1000, SharpMask yields slightly better AR than us. Note, however, that, in practice, it is not always possible to handle 1000 proposals in later processing stages, and many instance-level segmentation methods only consider 100 or 300, which is the regime where our approach performs best. In Fig.~\ref{fig:voc_2012_val_recall}, we report recall vs IOU threshold for all methods. Interestingly, even for 1000 segmentation proposals, our results outperform most of the baselines at high IOU thresholds. We refer the reader to the supplementary material for a comparison of the methods across different object sizes.

\begin{table}[!t]
\centering
\resizebox{0.8\linewidth}{!}{%
\begin{tabular}{rccc}
\hline \Tstrut
PASCAL VOC 2012 						& AR@10 	& AR@100 	& AR@1000	\\ \hline \Tstrut
Selective Search ~\cite{Uijlings13} 	& 7.0 		& 23.5 		& 43.3 		\\
MCG ~\cite{Arbelaez_2014_CVPR}			& 18.9 		& 36.8 		& 49.5 		\\
Deep-Mask ~\cite{DeepMask}				& 30.3 		& 45.0 		& 52.6 		\\
Sharp-Mask ~\cite{SharpMask}			& 33.3 		& 48.8 		& $\B{56.5}$ \\
MNC ~\cite{Jifeng15cocoseg} 			& 33.4 		& 48.5 		& 53.8 		\\
InstanceFCN ~\cite{Jifeng16cocoseg}		& 38.9 		& 49.7 		& 52.6 		\\ 
MNC+score 								& 45.7 		& 49.1 		& 52.5 		\\ \hline \Tstrut
OMN (ours) 								& $\B{47.8}$ & $\B{51.8}$ & 54.7 	\\ \hline
\end{tabular}} \vspace{1.0ex}
\caption{\textbf{Evaluation of our OMN on the PASCAL VOC 2012 validation set.} We compare our method with state-of-the-art segmentation proposal baselines according to the criteria of~\cite{HariharanAGM14, DBLP:COCO14}. Note that our approach outperforms the state-of-the-art methods for the top 10 and 100 segmentation proposals, which correspond to the most common scenarios when later processing is involved, e.g., instance level segmentation.}
\label{tab:voc_2012_val_table} \vspace{-3mm}
\end{table}

\section{Conclusion}

In this paper, we have introduced a distance transform-based mask representation that allows us to predict instance segmentations beyond the limits of initial bounding boxes. We have then shown how to infer and decode this representation with a fully-differentiable Object Mask Network (OMN) relying on a residual-deconvolutional architecture. We have then employed this OMN to develop a Boundary-Aware Instance Segmentation (BAIS) network. Our experiments on Pascal VOC 2012 and Cityscapes have demonstrated that our BAIS network outperforms the state-of-the-art instance-level semantic segmentation methods. In the future, we intend to replace the VGG16 network we rely on with deeper architectures, such as residual networks, to further improve the accuracy of our framework.

{\small
\bibliographystyle{ieee}
\bibliography{egbib}

\begin{thebibliography}{10}\itemsep=-1pt

\bibitem{Arbelaez_2014_CVPR}
P.~Arbelaez, J.~Pont-Tuset, J.~T. Barron, F.~Marques, and J.~Malik.
\newblock Multiscale combinatorial grouping.
\newblock In {\em CVPR}, 2014.

\bibitem{BaiU16}
M.~Bai and R.~Urtasun.
\newblock Deep watershed transform for instance segmentation.
\newblock In {\em CVPR}, 2017.

\bibitem{borgefors1986distance}
G.~Borgefors.
\newblock Distance transformations in digital images.
\newblock {\em Computer vision, graphics and image processing}, 1986.

\bibitem{chen2014semantic}
L.~C. Chen, G.~Papandreou, I.~Kokkinos, K.~Murphy, and A.~L. Yuille.
\newblock Semantic image segmentation with deep convolutional nets and fully
  connected crfs.
\newblock In {\em ICLR}, 2015.

\bibitem{Cordts2016Cityscapes}
M.~Cordts, M.~Omran, S.~Ramos, T.~Rehfeld, M.~Enzweiler, R.~Benenson,
  U.~Franke, S.~Roth, and B.~Schiele.
\newblock The cityscapes dataset for semantic urban scene understanding.
\newblock In {\em CVPR}, 2016.

\bibitem{Jifeng16cocoseg}
J.~Dai, K.~He, Y.~Li, S.~Ren, and J.~Sun.
\newblock Instance-sensitive fully convolutional networks.
\newblock In {\em ECCV}, 2016.

\bibitem{DaiH015}
J.~Dai, K.~He, and J.~Sun.
\newblock Convolutional feature masking for joint object and stuff
  segmentation.
\newblock In {\em CVPR}, 2015.

\bibitem{Jifeng15cocoseg}
J.~Dai, K.~He, and J.~Sun.
\newblock Instance-aware semantic segmentation via multi-task network cascades.
\newblock In {\em CVPR}, 2016.

\bibitem{Everingham12}
M.~Everingham, L.~Van~Gool, C.~K.~I. Williams, J.~Winn, and A.~Zisserman.
\newblock The {PASCAL} {V}isual {O}bject {C}lasses {C}hallenge 2012 {(VOC2012)}
  {R}esults.
\newblock
  http://www.pascal-network.org/challenges/VOC/voc2012/workshop/index.html.

\bibitem{farabet2013learning}
C.~Farabet, C.~Couprie, L.~Najman, and Y.~LeCun.
\newblock Learning hierarchical features for scene labeling.
\newblock {\em IEEE TPAMI}, 2013.

\bibitem{Girshick15}
R.~Girshick.
\newblock Fast r-cnn.
\newblock In {\em ICCV}, 2015.

\bibitem{gupta2014learning}
S.~Gupta, R.~Girshick, P.~Arbel{\'a}ez, and J.~Malik.
\newblock Learning rich features from rgb-d images for object detection and
  segmentation.
\newblock In {\em ECCV}, 2014.

\bibitem{BharathICCV2011}
B.~Hariharan, P.~Arbelaez, L.~Bourdev, S.~Maji, and J.~Malik.
\newblock Semantic contours from inverse detectors.
\newblock In {\em ICCV}, 2011.

\bibitem{HariharanAGM14}
B.~Hariharan, P.~Arbel\'{a}ez, R.~Girshick, and J.~Malik.
\newblock Simultaneous detection and segmentation.
\newblock In {\em ECCV}, 2014.

\bibitem{HariharanAGM15}
B.~Hariharan, P.~Arbel\'{a}ez, R.~Girshick, and J.~Malik.
\newblock Hypercolumns for object segmentation and fine-grained localization.
\newblock In {\em CVPR}, 2015.

\bibitem{he2014exemplar}
X.~He and S.~Gould.
\newblock An exemplar-based crf for multi-instance object segmentation.
\newblock In {\em CVPR}, 2014.

\bibitem{JanvandenBrand2016}
R.~M. Jan van~den Brand, Matthias~Ochs.
\newblock Instance-level segmentation of vehicles using deep contours.
\newblock In {\em Workshop on Computer Vision Technologies for Smart Vehicle,
  in ACCV}, 2016.

\bibitem{kimmel1996sub}
R.~Kimmel, N.~Kiryati, and A.~M. Bruckstein.
\newblock Sub-pixel distance maps and weighted distance transforms.
\newblock {\em Journal of Mathematical Imaging and Vision}, 1996.

\bibitem{kirillov-2016-arxiv}
A.~Kirillov, E.~Levinkov, B.~Andres, B.~Savchynskyy, and C.~Rother.
\newblock {InstanceCut}: from edges to instances with multicut.
\newblock In {\em CVPR}, 2017.

\bibitem{DBLP:conf/eccv/KrahenbuhlK14}
P.~Kr{\"{a}}henb{\"{u}}hl and V.~Koltun.
\newblock Geodesic object proposals.
\newblock In {\em ECCV}, 2014.

\bibitem{LiHM15}
K.~Li, B.~Hariharan, and J.~Malik.
\newblock Iterative instance segmentation.
\newblock In {\em CVPR}, 2016.

\bibitem{LiangArxiv15}
X.~Liang, Y.~Wei, X.~Shen, J.~Yang, L.~Lin, and S.~Yan.
\newblock Proposal-free network for instance-level object segmentation.
\newblock {\em CoRR}, abs/1509.02636, 2015.

\bibitem{DBLP:COCO14}
T.~Lin, M.~Maire, S.~Belongie, L.~D. Bourdev, R.~B. Girshick, J.~Hays,
  P.~Perona, D.~Ramanan, P.~Doll{\'{a}}r, and C.~L. Zitnick.
\newblock Microsoft {COCO:} common objects in context.
\newblock {\em CoRR}, abs/1405.0312, 2014.

\bibitem{FCN}
J.~Long, E.~Shelhamer, and T.~Darrell.
\newblock Fully convolutional networks for semantic segmentation.
\newblock In {\em CVPR}, 2015.

\bibitem{DeepMask}
P.~O. Pinheiro, R.~Collobert, and P.~Dollár.
\newblock Learning to segment object candidates.
\newblock In {\em NIPS}, 2015.

\bibitem{SharpMask}
P.~O. Pinheiro, T.-Y. Lin, R.~Collobert, and P.~Dollár.
\newblock Learning to refine object segments.
\newblock In {\em ECCV}, 2016.

\bibitem{RecAttend}
M.~Ren and R.~S. Zemel.
\newblock End-to-end instance segmentation and counting with recurrent
  attention.
\newblock In {\em CVPR}, 2017.

\bibitem{fasterRCNN}
S.~Ren, K.~He, R.~Girshick, and J.~Sun.
\newblock Faster r-cnn: Towards real-time object detection with region proposal
  networks.
\newblock In {\em NIPS}, 2015.

\bibitem{Romera-ParedesT15}
B.~Romera{-}Paredes and P.~H.~S. Torr.
\newblock Recurrent instance segmentation.
\newblock In {\em ECCV}, 2016.

\bibitem{scharr2014annotated}
H.~Scharr, M.~Minervini, A.~Fischbach, and S.~A. Tsaftaris.
\newblock Annotated image datasets of rosette plants.
\newblock In {\em ECCV}, 2014.

\bibitem{simonyan2014very}
K.~Simonyan and A.~Zisserman.
\newblock Very deep convolutional networks for large-scale image recognition.
\newblock {\em CoRR}, abs/1409.1556, 2014.

\bibitem{TigheNL14}
J.~Tighe, M.~Niethammer, and S.~Lazebnik.
\newblock Scene parsing with object instances and occlusion ordering.
\newblock In {\em CVPR}, 2014.

\bibitem{UhrigCFB16}
J.~Uhrig, M.~Cordts, U.~Franke, and T.~Brox.
\newblock Pixel-level encoding and depth layering for instance-level semantic
  labeling.
\newblock In {\em GCPR}, 2016.

\bibitem{Uijlings13}
J.~Uijlings, K.~van~de Sande, T.~Gevers, and A.~Smeulders.
\newblock Selective search for object recognition.
\newblock {\em IJCV}, 2013.

\bibitem{Ziyu15}
Z.~Zhang, S.~Fidler, and R.~Urtasun.
\newblock {Instance-Level Segmentation with Deep Densely Connected MRFs}.
\newblock In {\em CVPR}, 2016.

\bibitem{ZhangSFU15}
Z.~Zhang, A.~Schwing, S.~Fidler, and R.~Urtasun.
\newblock Monocular object instance segmentation and depth ordering with cnns.
\newblock In {\em ICCV}, 2015.

\end{thebibliography}
}

\end{document}